\documentclass[final,3p,times,twocolumn]{elsarticle}
\usepackage{amssymb,amsmath}
\usepackage{amsthm}
\usepackage[ngerman, english]{babel}
\usepackage{url}            % simple URL typesetting
\usepackage{amsfonts}       % blackboard math symbols
\usepackage{xcolor}         % colors

\newcommand{\R}{\mathbb{R}}

\newcommand{\Ical}{\mathcal{I}}
\newcommand{\Ncal}{\mathcal{N}}
\newcommand{\Ucal}{\mathcal{U}}
\newcommand{\Ycal}{\mathcal{Y}}

\newcommand{\norm}[1]{\left\lVert#1\right\rVert}
\newcommand{\abs}[1]{\left\vert#1\right\vert}

\newcommand{\revision}[1]{\textcolor{black}{#1}}

\newcommand{\dint}[1]{\,\mathrm{d}#1}
\newcommand{\pder}[2]{\frac{\partial #1}{\partial #2}}
\newcommand{\pderHigher}[3]{\frac{\partial^{#3} #1}{\partial #2^{#3}}}
\newcommand{\dt}{\dint{t}}

\newtheorem{remark}{Remark}

\journal{Physica D}

\begin{document}

\begin{frontmatter}



\title{Distributed Control of Partial Differential Equations Using Convolutional Reinforcement Learning}

 \author[label1]{Sebastian Peitz}
 \author[label1]{Jan Stenner}
 \author[label1]{Vikas Chidananda}
 \affiliation[label1]{organization={Department of Computer Science, Paderborn University},
             city={Paderborn},
             country={Germany}}
             
\author[label2]{Oliver Wallscheid}
 \affiliation[label2]{organization={Department of Electrical Engineering, Paderborn University},
             city={Paderborn},
             country={Germany}}

\author[label3]{Steven L.\ Brunton}
 \affiliation[label3]{organization={Department of Mechanical Engineering, University of Washington},
             city={Seattle},
             state={WA},
             country={USA}}
             
\author[label4]{Kunihiko Taira}
 \affiliation[label4]{organization={Department of Mechanical and Aerospace Engineering, University of California},
             city={Los Angeles},
             state={CA},
             country={USA}}

\begin{abstract}
We present a convolutional framework which significantly reduces the complexity and thus, the computational effort for distributed reinforcement learning control of dynamical systems governed by partial differential equations (PDEs). Exploiting translational equivariances, the high-dimensional distributed control problem can be transformed into a multi-agent control problem with many identical, uncoupled agents. Furthermore, using the fact that information is transported with finite velocity in many cases, the dimension of the agents' environment can be drastically reduced using a convolution operation over the state space of the PDE, \revision{by which we effectively tackle the curse of dimensionality otherwise present in deep reinforcement learning}. In this setting, the complexity can be flexibly adjusted via the kernel width or by using a stride greater than one (\revision{meaning that we do not place an actuator at each sensor location}). Moreover, scaling from smaller to larger \revision{domains} -- or the transfer between different domains -- becomes a straightforward task requiring little effort. We demonstrate the performance of the proposed framework using several PDE examples with increasing complexity, where stabilization is achieved by training a low-dimensional deep deterministic policy gradient agent using minimal computing resources.
\end{abstract}

% \begin{highlights}
% \item We present a framework which significantly reduces the computational effort for distributed reinforcement learning control of dynamical systems governed by partial differential equations (PDEs).
% \item Using a convolution preprocessing step, we transform the control task into a multi-agent problem with agents that have significantly reduced state and action space dimensions. 
% \item By exploiting symmetries (translational equivariance), all agents are statistically identical -- meaning that on average, they observe the same states and perform the same actions. It thus suffices to train a single agent and deploy identical copies.
% \item Our approach avoids the curse of dimensionality, while at the same time allowing for plug-and-play transfer between different domains.
% \item Due to these reductions, training can be realized within just a few minutes on a consumer grade laptop.
% \end{highlights}


\end{frontmatter}


\section{Introduction}
Distributed control of dynamical systems governed by partial differential equations (PDEs) is a challenging task -- both from a control theoretical as well as a computational point of view -- with numerous important applications, such as the control of chemical processes~\cite{Chr01}, turbulence control~\cite{BMT01}, and robotic systems~\cite{ZXNC05}.
Due to the dependency on both space and time, these control problems exhibit a very large number of degrees of freedom (DOF) for both the system state as well as the control input, which in particular renders real-time control difficult. Even offline computations can quickly become prohibitively expensive, and in terms of machine learning control, the large number of DOFs calls for an extremely large number of trainable parameters and substantial requirements regarding the training data, rendering learning expensive in terms of data and computation~\cite{BNK20}.

A plethora of data-driven and machine-learning approaches for PDE-constrained control have been proposed in recent years, most prominently in the field of fluid mechanics~\cite{DBN17}. Among these, \emph{reinforcement learning} (RL) plays an increasingly important role~\cite{NLK21,KIG22,VSA+22,VLLR22,WYH+22,WP23}.
However, the complexity of the dynamics and the very large number of parameters call for additional measures, by means of exploiting patterns within the dynamics or explicitly including system knowledge. 
In the first case, one can derive data-driven surrogate models of lower dimension using, e.g., the \emph{Proper Orthogonal Decomposition}~\cite{KV99}, \emph{Dynamic Mode Decomposition}~\cite{PBK15,KM18,PK19,POR20} or models based on \emph{deep neural networks}~\cite{RKJ+19,BPB+20}.
In the latter case, we may make use of the governing equations in the form of \emph{Physics-Informed Neural Networks}~\cite{RPK19}, or reduce the complexity by exploiting symmetries such as translational or rotational invariances. This can be useful both in the control setting (see, for instance,~\cite{OP21} for \emph{model predictive control} of ordinary differential equations with symmetries) as well as for prediction, see~\cite{PHG+18} for chaotic PDEs. In the latter paper, the authors also exploit the fact that mass and energy are transported with finite velocity, which allows them to replace a global surrogate model by a set of identical, locally coupled agents. Similar considerations also allow for the construction of network models for the analysis of complex systems such as turbulent flows~\cite{TNB16}.

In the RL literature, multi-agent approaches have until now mainly been used for distributed control of interconnected systems \cite{BBS08,CSA20,NJ18}, mostly focusing on individual agents and their interaction and less on training and data efficiency. The case of identical agents was addressed in~\cite{NLK21} and~\cite{BK22} in terms of enhancing turbulence modeling for wall-bounded flows via RL, but not for control. 
\revision{
The closest relation in terms of exploiting invariances is \cite{BRV+19} in a multi-agent context. Other works have appeared very recently, where such a multi-agent concept has been successfully used as well, see. e.g., \cite{GRS+23,SLIH23} for drag reduction in turbulent channel flows, or \cite{VRV+23} for Rayleigh-Bénard convection.
The main difference is that we formalize our approach in terms of convolutions, which is very general. Moreover, we also address the straightforward transfer of trained models to other domains, which we consider a key component towards a more flexible application of RL in engineering applications.}

\begin{figure}[t]
	\centering
	\includegraphics[width=\columnwidth]{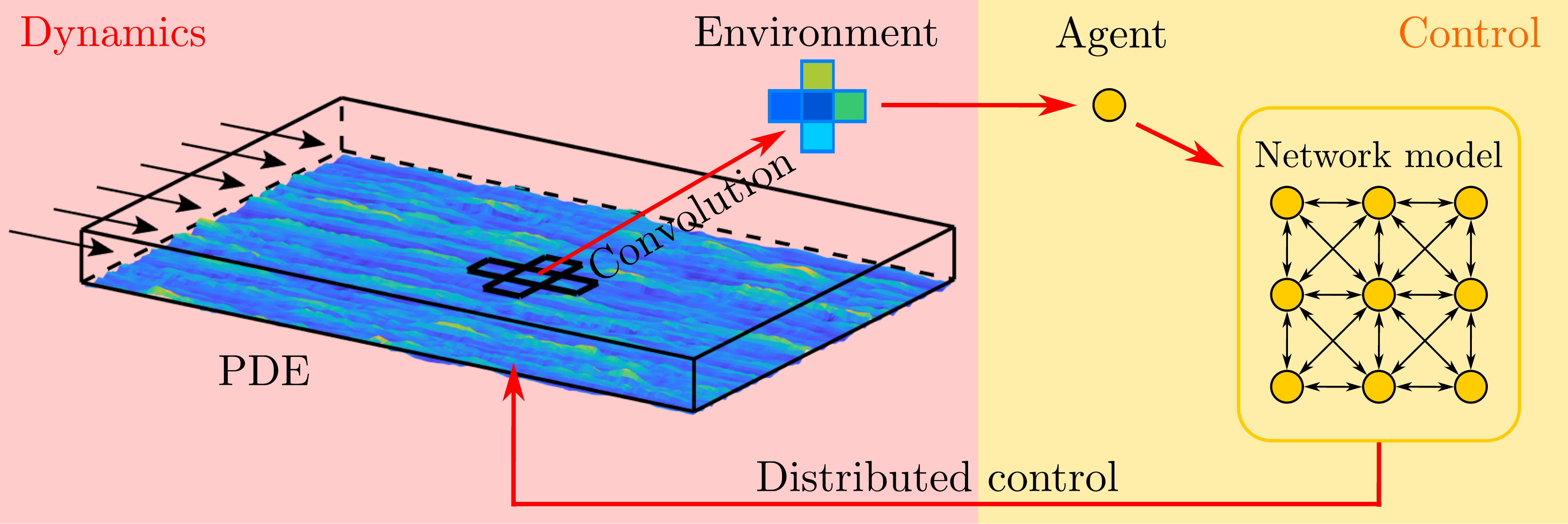}
	\caption{The convolutional reinforcement learning framework. The coupling in the network model on the right refers only to shared state information; the decision making process is entirely decoupled.}
	\label{fig:CRL}
\end{figure}

Our approach to the distributed control problem is via a multi-agent RL approach of uncoupled, identical agents (see Fig.\ \ref{fig:CRL}) with the details presented in Section \ref{sec:CRL}. 
The locality and identity of the agents is ensured by applying a convolution operation to the PDE state in order to obtain a spatially confined state for the RL agent. \revision{The usage of a mostly local reward allows each agent to learn individually and to choose its actions based on local information only. Finally, translational equivariance ensures that all agents are statistically identical such that we only need to train a single agent, of which we can then deploy as many copies as we desire.}
The advantages are 
\begin{itemize}
    \item a massive reduction of the agent's state and action space dimensions, which allows for much smaller agents \revision{and thus tackles the curse of dimensionality},
    \item a strong increase in the available training data, as all agents share the same parameters and training data set,
    \item the transfer of trained agents  to various spatial domains in a simple plug-and-play manner.
\end{itemize}
%  as well as 
% In addition, 
We observe in several examples (Section \ref{sec:Examples}) that we can train a stabilizing feedback controller using very limited computational resources. For instance, we can learn a stabilizing policy for the Kuramoto--Sivashinsky equation on a large domain of size $L=500$ with $P=200$ actuators within just a few minutes on a consumer grade laptop, \revision{which is a significant increase by a factor of more than $20$ over other state-of-the-art control schemes, where a domain size of $L=22$ (the onset of chaos) is considered at most.}

%We note that this work is closely related to the literature on multi-agent reinforcement learning, see, e.g.,~\cite{BBS08} for a survey,~\cite{CSA20} for systems with shared experience, or~\cite{NJ18} for distributed control of interconnected systems. However, these works mainly focus on individual agents and their interaction, and less on training and data efficiency. The case of identical agents was addressed in~\cite{NLK21} and~\cite{BK22} in terms of enhancing turbulence modeling for wall-bounded flows via RL, but not for control.

%%%%%%%%%%%%%%%%%%%%%%%%%%%%%%%%%%%%%%%%%%%%%%%%%%%%%%%%%%%%%%%%%%%%%%%%%%%%%%%%%%%%%%%%%%%%%%
\section{Reinforcement learning}
\label{sec:RL}
We here only give a very brief overview of reinforcement learning (RL); a much more detailed introduction can be found in, e.g.,~\cite{SB18}.
The standard reinforcement learning setup consists of an \emph{agent} interacting with an \emph{environment} (stochastic or deterministic) in discrete time steps. At each time step $k$, the agent receives an \emph{observation} $\tilde{y}_k\in\tilde{\Ycal}$, takes an \emph{action} $\tilde{u}_k\in\Ucal$ and receives a \emph{reward} $R_k\in\R$. The environment may be only partially observed, and the observation may also consist of delay coordinates. 

The agent's behavior is defined by a \emph{policy} $\pi$, which is a mapping from the state space to a probability distribution over the actions $\pi:\Ycal \rightarrow P(\Ucal)$. The environment dynamics is described by a \emph{Markov decision process (MDP)} with a state space $\Ycal$, action space $\Ucal\subseteq\R^m$, an initial state distribution $p(\tilde{y}_0)$, transition dynamics $p(\tilde{y}_{k+1} | \tilde{y}_k, \tilde{u}_k)$ and a reward function $r(\tilde{y}_k, \tilde{u}_k)$. 
The \emph{return} from a state, given a policy $\pi$, is defined as the sum of discounted future rewards 
$G= \mathbb{E}_{\pi} \left[\sum_{i=k}^p \gamma^{i-k} R_i\right]$, 
%\[
%    % R_k= \mathbb{E}_{\pi} \left[\sum_{i=k}^p \gamma^{i-k} r(\tilde{y}_k, \tilde{u}_k)\right], 
%    G= \mathbb{E}_{\pi} \left[\sum_{i=k}^p \gamma^{i-k} R_i\right], 
%\]
where $\gamma\in[0,1]$ is the discount factor. %, and $\tilde{Y}$ and $\tilde{U}$ are the random variables describing the state and input at iteration $k$, of which $\tilde{y}$ and $\tilde{u}_k$ are then realizations.
The \emph{goal} in reinforcement learning is now to learn the policy that maximizes the expected return, which is intricately related to Bellman's principle of optimality and the identification of the associated value function. 

As the central goal of this paper is to introduce a novel learning architecture which significantly simplifies the training by exploiting physical properties of the system, we will not focus on a specialized RL architecture. Instead, the method we use to train the individual agents is the well-known \emph{Deep Deterministic Policy Gradient (DDPG)} \cite{LHP+15} in its standard version (i.e., the implementation in the \emph{Julia} package \emph{ReinforcementLearning.jl}).
DDPG is a model-free \emph{actor-critic} algorithm based on the deterministic policy gradient~\cite{SLH+14} that can operate over continuous action spaces. Actor critic refers to a family of algorithms in which two networks are trained. The actor approximates the policy and decides which action to take, whereas the critic assesses the quality of the action taken, i.e., it approximates the value function.
Both actor and critic are approximated using deep feed-forward neural networks, and a second set of \emph{target networks} are used, which are responsible for the interactions with the environment and which are updated periodically from the former set of networks which are continuously trained from replay buffers. 
%The training process in RL is strongly motivated by trial and error, meaning that we \emph{explore} different actions and assess whether they yield a high reward or not. In the long run, we then \emph{exploit}, meaning that given a good approximation of the value function, we can choose actions that maximize the expected return.
%As we are simply using DDPG in its standard form and focusing on the task of exploiting system knowledge, we will not go into further details, but refer interested readers to the mentioned references for details.

%%%%%%%%%%%%%%%%%%%%%%%%%%%%%%%%%%%%%%%%%%%%%%%%%%%%%%%%%%%%%%%%%%%%%%%%%%%%%%%%%%%%%%%%%%%%%%
\section{Convolutional reinforcement learning}
\label{sec:CRL}

Our objective is to solve a distributed control problem governed by a partial differential equation (PDE).
The state $y:\Omega \times [0,T] \rightarrow \R^n$ is a function of time $t\in[0,T]$ and space $x\in\Omega$, and the dynamics is described by a nonlinear partial differential operator $\Ncal$, i.e., 
\[
	\pder{y}{t} = \Ncal(y,u), 
\]
with $u:\Omega \times [0,T] \rightarrow \R^m$ being the \emph{spatially distributed, time-dependent control input}. %, and the subscript $t$ denotes the partial derivative with respect to time. 
Throughout the paper, we assume that the dynamics are \revision{equivariant (or at least almost equivariant)} under translation, which is the case for all systems where the position $x$ does not explicitly appear in the PDE or in an inhomogeneous source term. %, nor some sort of inhomogeneous source term.

In order to deal with the discrete-time nature of RL, we directly introduce a \revision{zero order hold on the control and a} partial discretization in time
%\footnote{If we additionally use a spatial discretization (such as finite elements), we obtain a high-dimensional ordinary differential equation.} 
with constant time step $\Delta t = t_{k+1} - t_k$, $k=0,1,\ldots,p$, i.e.,
\[
	\Phi(y_k, u_k) = y_k + \int_{t_k}^{t_{k+1}} \Ncal(y(\cdot,t), u_k) \dt = y_{k+1},
\]
where $u(\cdot,t)=u_k$ is constant over the interval $[t_k, t_{k+1})$.
Using the above considerations, the control task can be formalized in an optimal control problem of the following form:
% \begin{equation} 
	\begin{align}
		\min_{u} J(y,u) &= \min_{u} \sum_{k=0}^{p} \ell\left(y_k,u_k\right) \label{eq:OCP}\\
		\mbox{s.t.} \quad y_{k+1} &= \Phi(y_k, u_k), \qquad k = 0,1,2,\ldots,p-1, \notag
	\end{align}
% \end{equation}
where $J$ is the objective functional over the time horizon $T = p\Delta t$, and $\ell$ is the \emph{stage cost}, e.g., a tracking term
\begin{align*}
	\ell\left(y_k,u_k\right) &= \norm{y_k - y_k^{\mathsf{ref}}}_{L^2(\Omega)}^2 + \lambda \norm{u_k}_{L^2(\Omega)}^2 \\ &= \int_\Omega \left(y_k(x) - y_k^{\mathsf{ref}}(x)\right)^2 + \lambda \left(u_k(x) \right)^2 \dint{x},
\end{align*}
where the regularization parameter $\lambda > 0$ penalizes the control cost.
In terms of RL, the stage cost $\ell$ can be seen as the negative reward.
Note that at each discrete point in time $k\Delta t$, the corresponding $u_k$ is still a function of space. %, i.e., it is infinite-dimensional. 

\subsection{Transformation into a multi-agent control problem}
We now introduce a convolution operator $ \psi$ into the objective functional $\ell$, which introduces a new spatial variable $c$ (e.g., the center of a Gaussian; for a detailed introduction to convolutional neural networks (CNNs), see~\cite{GBC16}):
\begin{equation}\label{eq:ConvP}
    \begin{aligned}
	\hat{\ell}_c\left(y_k,u_k\right) = &\int_\Omega \left[  \psi(x-c) \left(y_k(x) - y_k^{\mathsf{ref}}(x)\right) \right]^2 \\
    &+ \lambda \left[  \psi(x-c) u_k(x) \right]^2 \dint{x}.
    \end{aligned}
\end{equation}

\begin{figure}[t]
	\centering
	\includegraphics[width=\columnwidth]{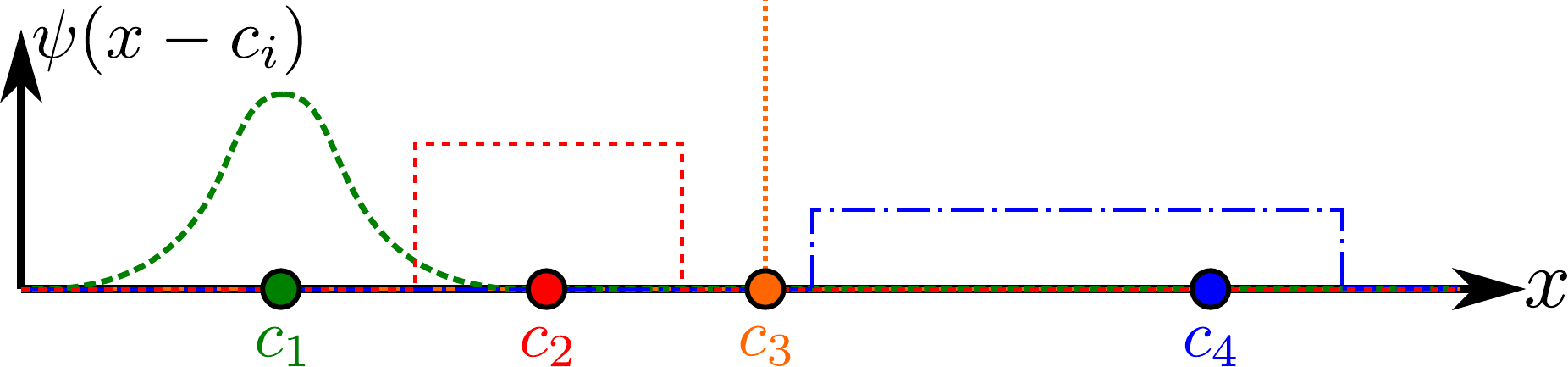}
	\caption{\revision{Examples of 1D convolution kernels.}}
	\label{fig:Conv1D}
\end{figure}

\begin{remark}
The convolution operation in \eqref{eq:ConvP} can be realized in many ways in practice, \revision{see Fig.\ \ref{fig:Conv1D} for a few examples in the case of one-dimensional spatial domains (higher-dimensional cases are analogous)}. 
For instance, we can use Gaussians, but also discontinuous kernels which are simpler to use for spatially discretized systems. In the discretized setting, the kernel cells may contain the values of individual grid nodes, but also pooled values such as spatial averages (see Fig.\ \ref{fig:CRL}).
Finally, we may also consider additional knowledge (such as dominant directions) by introducing non-symmetric kernels. \revision{It should be noted that the choice of this kernel also has an impact on the control authority one has over the system of interest.}
\end{remark}

We can now follow the standard approach for CNNs and define discrete convolution kernels with multiple cells (such as the one indicated in Fig.\ \ref{fig:CRL}) that move over the state space with a fixed stride. %, see~\cite{GBC16} (Ch.\ 9) for details. 
This yields a set of $M$ convolutions $ \psi$ with centers $c_1,\ldots,c_M$, which allows us to split the integral over the domain $\Omega$ in the objective functional $\ell$ of problem \eqref{eq:OCP}. If the kernels are indicator functions %(as illustrated in Fig.\ \ref{fig:Conv1D} in blue and red) 
with disjoint support, then we obtain
\begin{equation*}%\label{eq:summation}
    \begin{aligned}
	\int_\Omega 1 \dint{x} &= \sum_{i=1}^M \int_\Omega  \psi(x-c_i) \dint{x} \\ \Rightarrow \qquad \sum_{i=1}^M \hat{\ell}_{c_i}\left(y,u\right) &= \ell\left(y,u\right).
    \end{aligned}
\end{equation*}
%where we have introduced $y_i=y(x_i)$ and $u_i=u(x_i)$ for brevity.
If the individual convolution operators overlap -- e.g., when using a $3$-dimensional convolution kernel with stride 1 -- then the summation yields the value of $\ell$ multiplied by a scalar.

Applying the convolution for each kernel center results in $n \times M$ (convolved) state variables and $m \times M$ optimization parameters at each time instance $k$, %($u\in\R^{m\times M}$) 
\begin{align*}
	\tilde{y}_{k,i} &= \int_\Omega  \psi(x-c_i) y_k(x) \dint{x}, \quad i=1,\ldots,M \\ \mbox{and} \quad
	\tilde{u}_{k,i} &= \int_\Omega  \psi(x-c_i) u_k(x) \dint{x}, \quad i=1,\ldots,M.
\end{align*}
Based on this convolution, we can now formulate a control problem closely related to \eqref{eq:OCP} in the new variables $\tilde{y}$ and $\tilde{u}$.
Using the above summation for the objective function and introducing individual dynamics $\tilde\Phi_i$ for the $M$ subsystems located at $c_1,\ldots,c_M$, we obtain the network control problem
\begin{equation} \label{eq:NetOCP}
	\begin{aligned}
		&\min_{\tilde{u}\in \R^{M \times p}}\sum_{k=0}^{p} \sum_{i=1}^M \hat{\ell}_{c_i}\left(\tilde{y}_{k,i}, \tilde{u}_{k,i}\right) \\
		\mbox{s.t.} \quad &\tilde{y}_{k+1,i} = \tilde\Phi_i(\tilde{y}_k, \tilde{u}_k), \qquad \begin{array}{l} k = 0,\ldots,p-1 \\ \,i = 1,\ldots,M \end{array}.
	\end{aligned}
\end{equation}
\begin{remark}\label{rem:diracKernel}
    If we use Dirac delta functions for $\psi$, then problem \eqref{eq:NetOCP} is equivalent to the spatial discretization of problem \eqref{eq:OCP} using finite differences. For other kernels, this can still be interpreted as a Galerkin-type discretization in space.
\end{remark}
\noindent
Note that the right-hand sides of the $M$ systems $\tilde\Phi_1,\ldots,\tilde\Phi_M$ still depend on the state and control of all systems such that we have a fully connected network. 
\revision{Thus, in order to split the optimal control problem \eqref{eq:NetOCP} and to introduce a distributed computation of the optimal control $\tilde{u}_i$, we will pose additional assumptions as discussed in the next section.}
%We can not decouple the calculation of the $\tilde{u}_i$ unless making more assumptions, which will be discussed in the following section.

\subsection{Complexity reduction}
To obtain a significant simplification of problem \eqref{eq:NetOCP}, 
we exploit two important physical properties:
\begin{itemize}
    \item in many physical processes, information is transported with finite velocity,
    \item due to the translational equivariance in the dynamics, all local systems are identical, i.e., $\tilde\Phi_1 =\ldots=\tilde\Phi_M=\overline{\Phi}$.
\end{itemize}
The first point effectively means that the dynamics are local, to some extent, and state values at locations far away do not have an immediate impact on the local evolution of $y$; however, they may have an effect after finite time, so that the time step $\Delta t$ must be sufficiently small, analogous to the CFL condition in computational fluid dynamics~\cite{FP02}.
% (even though they obviously may have in the long run, such that the time step $\Delta t$ has to be sufficiently small, not unlike the CFL condition in computational fluid dynamics~\cite{FP02}).
The second point will allow us to obtain a training task with much lower dimension.
Both facts have been exploited in~\cite{PHG+18} for the efficient construction of predictive models for distributed systems using a network of surrogate models, each of which makes their prediction using local information only. 
For our control problem, this means that we introduce the assumption that the right-hand side of the dynamics $\tilde\Phi_i$ in problem \eqref{eq:NetOCP} no longer depends on the entire state $\tilde y$, but on a subset $\tilde y_{\Ical_i}$ localized around $c_i$, where $\Ical_i\subset\{1,\ldots,N\}$. 
%On the other hand, due to the translational invariance in the dynamics, all local systems are identical, i.e., $\tilde\Phi_1 =\ldots=\tilde\Phi_M=\overline{\Phi}$:
In combination with the fact that all local systems are identical, we obtain
\[
	\tilde{y}_{k+1,i} = \overline{\Phi}(\tilde{y}_{k,\Ical_i}, \tilde{u}_k), \qquad i = 1,\ldots,M.
\]
The entries considered in $\Ical_i$ are determined by the choice of the kernel. For instance, the kernel shown in Fig.\ \ref{fig:CRL} results in a five-dimensional set $\Ical_i$ consisting of $\tilde{y}_i$ as well as the direct neighbors in both spatial directions (i.e., the standard finite-difference stencil).

\begin{remark}
    Even more generally, the connectivity between the agents could also be encoded in a sparse \emph{connectivity matrix}. When considering neighboring agents only, this would result in a sparse multi-band matrix, but other patterns are possible as well. For instance, one could devise a connectivity pattern that is specifically tailored to the dynamics, i.e., to the interaction between different regions within the domain~\cite{TNB16}.
\end{remark}

\begin{figure*}[t]
	\centering
	%\includegraphics[width=.6\textwidth]{graphics/AgentKS.png}
% 	\caption{Data and parameter sharing in the convolutional RL framework. We have $M$ identical agents, each of which has a local environment of dimension $S$ (here, $S=5$) and takes a scalar control decision.}
	\includegraphics[width=.9\textwidth]{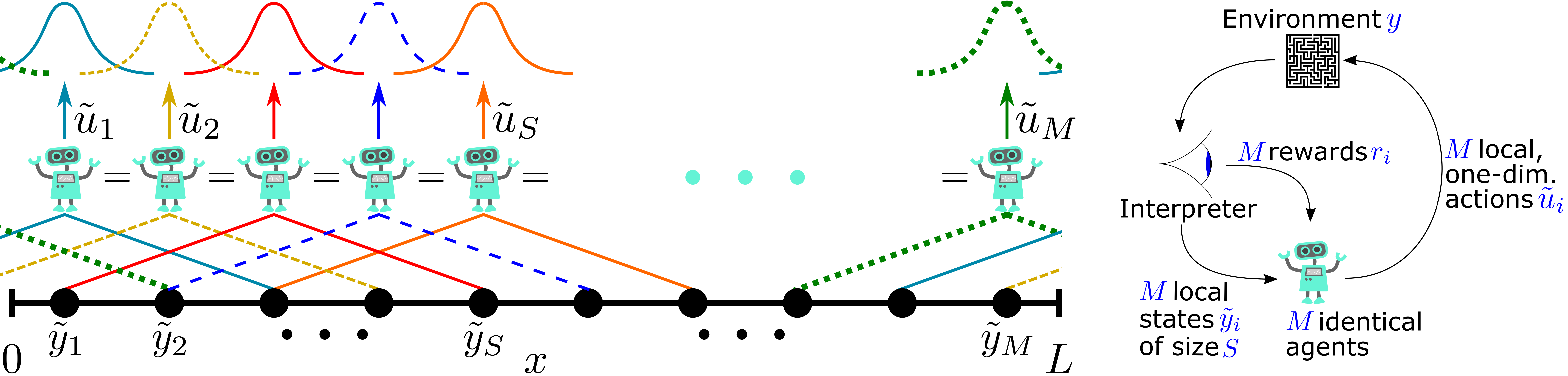}
	\caption{Data and parameter sharing in the convolutional RL framework (here for periodic boundary conditions). Left: We have $M$ identical agents, each of which has a local environment of dimension $S$ (here, $S=5$) and takes a scalar control decision. Right: Overall framework, in which we have $M$ times the data for training.}
	\label{fig:AgentKS}
\end{figure*}

\subsection{The reinforcement learning problem}
As a result of the discussion above, we obtain a sparsely connected network in problem \eqref{eq:NetOCP} instead of a fully connected one, which could in principle be solved using techniques from the area of \emph{cooperative control}, see, e.g.,~\cite{RC11,SCBM15}. However, (a) the control parameters $\tilde{u}_i$ are still not decoupled and (b) the longer the horizon $p$ in our control problem, the stronger the coupling effectively becomes. The latter is due to the fact that the states $\tilde{y}_i$ are part of multiple agents' environments such that the action an agent takes has also an impact on the neighboring environments. Thus, for high-dimensional PDEs, the problem is still very challenging. Finally, if we do not choose a Dirac kernel (see Remark \ref{rem:diracKernel}), then we cannot rely on finite differences as the model for $\overline{\Phi}$, hence we will not even have a computational model.
As a remedy, we add another assumption which is also based on the observation of locality. This assumption is that the control also only has a local influence over short time horizons, which means that we do not have to optimize over all inputs at the same time.

Following these considerations, we define the following RL framework: % in the sense of Section \ref{sec:RL}:
\begin{itemize}
	\item The agent's environment $\Ycal$ only consists of the states
	%and inputs 
	within the convolution kernel, i.e., $\Ycal_i = \{ \tilde{y}_{\Ical_i}\}$. 
%	\[
%	   % \Ycal_i = \{ \tilde{y}_{\Ical_i},\tilde{u}_{\Ical_i} \}. 
%	    \Ycal_i = \{ \tilde{y}_{\Ical_i}\}. 
%	\]
	Denoting the dimension of the kernel by $\abs{\Ical}=S$, we find that dim$(\Ycal_i)=n\times S \ll n\times M$.\\
%	\emph{Remark: One can in principle include additional information such as time delays.}
	\item The action space $\Ucal$ only consists of the input at the $i$-th cell, i.e., dim$(\Ucal)=m \ll m \times M$.
	\item The reward should mainly rely on local information, as this is required for assessing the agent's local performance. To a small extent, global contributions can be used to take unactuated areas into account, as well as to have an averaging effect over all spatial realizations of the agent:
	\[
		r_i(	\tilde{y}_{\Ical_i},\tilde{u}_{\Ical_i}) = - \sum_{j\in\Ical_i} \hat{\ell}\left(\tilde{y}_j,\tilde{u}_j\right) - \beta \ell\left(\tilde{y}_j,\tilde{u}_j\right),
	\]
	with some fixed value $\beta\geq 0$. We believe that the global information becomes more important if we cannot apply a control to the entire domain, but only a subset of $\Omega$. However, the aspect of global rewards is left to future investigations.
\end{itemize}
We comment on the specifics of these choices in the examples in Section \ref{sec:Examples}.

\begin{remark}[\revision{Difference to standard convolutions in CNNs}]\label{rem:ComparisonCNN}
\revision{We would like to emphasize that our approach is different from simply training a deep RL agent that contains a convolutional neural network. While it is true that we obtain parameter sharing in both cases, our convolution operation works as a \emph{preprocessing step}, by which the agent's state and action spaces are very small. If we simply used a CNN architecture, we would still have to deal with extremely high state and action space dimensions and thus still suffer from the curse of dimensionality. On the other hand, our convolution operation reduces the dimensions by such a large amount that we can use simple and very small feed-forward networks to approximate the policy.}
\end{remark}

\begin{remark}
    The question of the global reward can be crucial, not only in terms of finding a good policy, but also from a conceptual point of view. 
    %If we want to control parts of the domain that we can not see, then we do no longer have an MDP, but only a partially observable MDP (POMDP)~\cite{SB18}, which is considerably more challenging to control.
    \revision{Nevertheless, we believe that this assumption is \underline{much less restrictive in physics systems} than it would be, e.g., in the domain of computer games. This is due to the fact that many objectives such as stabilizing the dynamics, minimizing forces on surfaces, or maximizing the heat transfer, are integrals of local quantities over the domain of interest.}
\end{remark}

\subsection{Training using parameter and data sharing}

As the system is invariant under translation, we can use the identical agent at all $M$ locations, which has the two advantages that (a) we only have to train one model that is not even particularly large due to the small dimensions of the environment and action space, and (b) we can use the data acquired in all $M$ locations to train the same agent, meaning that we obtain $M$ rewards in every iteration. This concept is visualized in Figure \ref{fig:AgentKS}.

For training, we thus proceed according to the standard DDPG setting, only with $M$ identical agents. This means that we measure the environment by applying the convolution $M$ times, then decide on $M$ local actions independently and collect $M$ corresponding rewards. These $M$ additional training data samples are then fed to the DDPG replay buffer in order to continue training.

%%%%%%%%%%%%%%%%%%%%%%%%%%%%%%%%%%%%%%%%%%%%%%%%%%%%%%%%%%%%%%%%%%%%%%%%%%%%%%%%%%%%%%%%%%%%%%
\section{Examples}
\label{sec:Examples}
We evaluate the performance of the convolutional RL framework on different examples. For the Kuramoto--Sivashinsky equation, we first show that we can stabilize a chaotic PDE with small training effort. We further demonstrate that our approach scales well and that a transfer from one domain to the other is possible just as easily. Finally, we study the performance under nonhomogeneous disturbances.
In the next example, we study a more complicated control task for a two-component Chemotaxis system, where we control the cell density by varying the chemoattractant concentration using our framework.
For a two-dimensioal domain, we then study an isotropic turbulent flow.

\subsection{Kuramoto--Sivashinsky equation}

The first system we study is the 1D \emph{Kuramoto--Sivashinsky equation} (KS), which models the diffusive–thermal instabilities in a laminar flame front:
\begin{equation*}%\label{eq:KS}
    \begin{aligned}
        % y_t = -y y_x - y_{xx}-y_{xxxx} + \mu \cos\left(\frac{4 \pi x}{L}\right) + f(x,u),
        \pder{y}{t} = -y \pder{y}{x} - \pderHigher{y}{x}{2}-\pderHigher{y}{x}{4} + \mu \cos\left(\frac{4 \pi x}{L}\right) + f(x,u),
    \end{aligned}
\end{equation*}
with periodic boundary conditions and $\mu \geq 0$. %, and the subscript denotes derivatives w.r.t.\ time and space, respectively. 
Note that for $\mu>0$, this system is no longer invariant under translation in $x$. However, similar to~\cite{PHG+18}, we will see that our approach is capable of handling small disturbances of this kind.
The final term $f(x,u)$ is the control term:
\[
    f(x,u) = \sum_{i=1}^P u_i \psi(x - c_i), % \frac{\exp\left(-\frac{1}{2}(x-c_i)^2\right)}{b}
\]
where the control action $u$ is a scaling factor to $P$ different basis functions $\psi(x - c_i)$ located at $c_1,\ldots,c_P$. 
%These can be, for instance, the functions depicted in Figure \ref{fig:Conv1D}. 
For a radial basis function, we have 
%\[\psi(x - c_i)=\exp\left(-\frac{1}{2}\left(\frac{x-c_i}{\sigma}\right)^2\right),\] 
$\psi(x - c_i)=\exp\left(-\frac{1}{2}\left(\frac{x-c_i}{\sigma}\right)^2\right)$,
where $\sigma$ is an additional parameter determining the kernel width.
\begin{remark}
    Note that we have now separated the number of convolved states from the inputs, i.e., we do not have to use $P=M$. In many situations, we may have fewer actuators than sensors, which means $P<M$.
\end{remark}

\begin{table}[t]
    % \begin{adjustbox}{width=\columnwidth,center}
    \centering
    \caption{Numerical settings for the Kuramoto--Sivashinsky exmaple. In all cases, the kernel $\psi$ consists of Gaussians with $\sigma=0.8$.}
    \begin{tabular}{l|c|ccc}
        ~ & Var. & & Cases & \\ \hline
        Domain size & $L$ & 22 & 200 & 500 \\
        \# Sensors & $M$ & 8 & 80 & 200 \\
        \# Actuators & $P$ & 8 & 80 & 200 \\
        Agent state space dim. & $S$ & 1 & 1 & 1 \\
        \# Hidden l.\ Actor/Critic & ~ & 1/1 & 1/1 & 1/1 \\
        \# Neurons per layer & ~ & 6/140 & 6/140 & 6/140
    \end{tabular}
    % \end{adjustbox}
    \label{tab:KS}
\end{table}

For domains $\Omega=[0,L]$ with sufficiently large $L$, the dynamics exhibit chaotic behavior, rendering this a popular system for turbulence studies. Moreover, it has been studied in terms of control as well, both in theory~\cite{RY01} as well as in practice, see, e.g.,~\cite{BSA+19,LW21,ZG21} for reinforcement learning approaches. In all three articles, $P=4$ actuators are used on a domain of size $L=22$, which is comparatively small and is close to the onset of turbulence. 
In~\cite{LW21} and \cite{ZG21}, the goal is to minimize dissipation, which yields a global reward of the form 
$r = -\langle (y_{xx})^2 \rangle -\langle (y_x)^2 \rangle -\langle yf \rangle $,
% \[
%     r(t) = -\frac{1}{L} \int_{0}^L (y_{xx}(x,t))^2 + (y_x(x,t))^2 + y(x,t) f(x,u(t)) \dint{x}
% \]
%has to be maximized, 
where $\langle \cdot \rangle$ is the spatial average. In~\cite{BSA+19}, the task is to steer from one fixed point to another. As the final term $\langle yf \rangle$ is not invariant under translation, we here follow the latter and aim for the standard control task of driving $y$ towards zero. Introducing a regularization parameter $\alpha>0$, this yields
\[
    r = -\ell= -\left(\langle y^2 \rangle + \alpha \sum_{i=1}^P u_i^2 \left\langle \psi^2(x - c_i) \right\rangle\right).
\]

\begin{figure}[t!]
 	\centering
 	
 	\parbox[b]{\columnwidth}{
    \centering
	\includegraphics[width=\columnwidth]{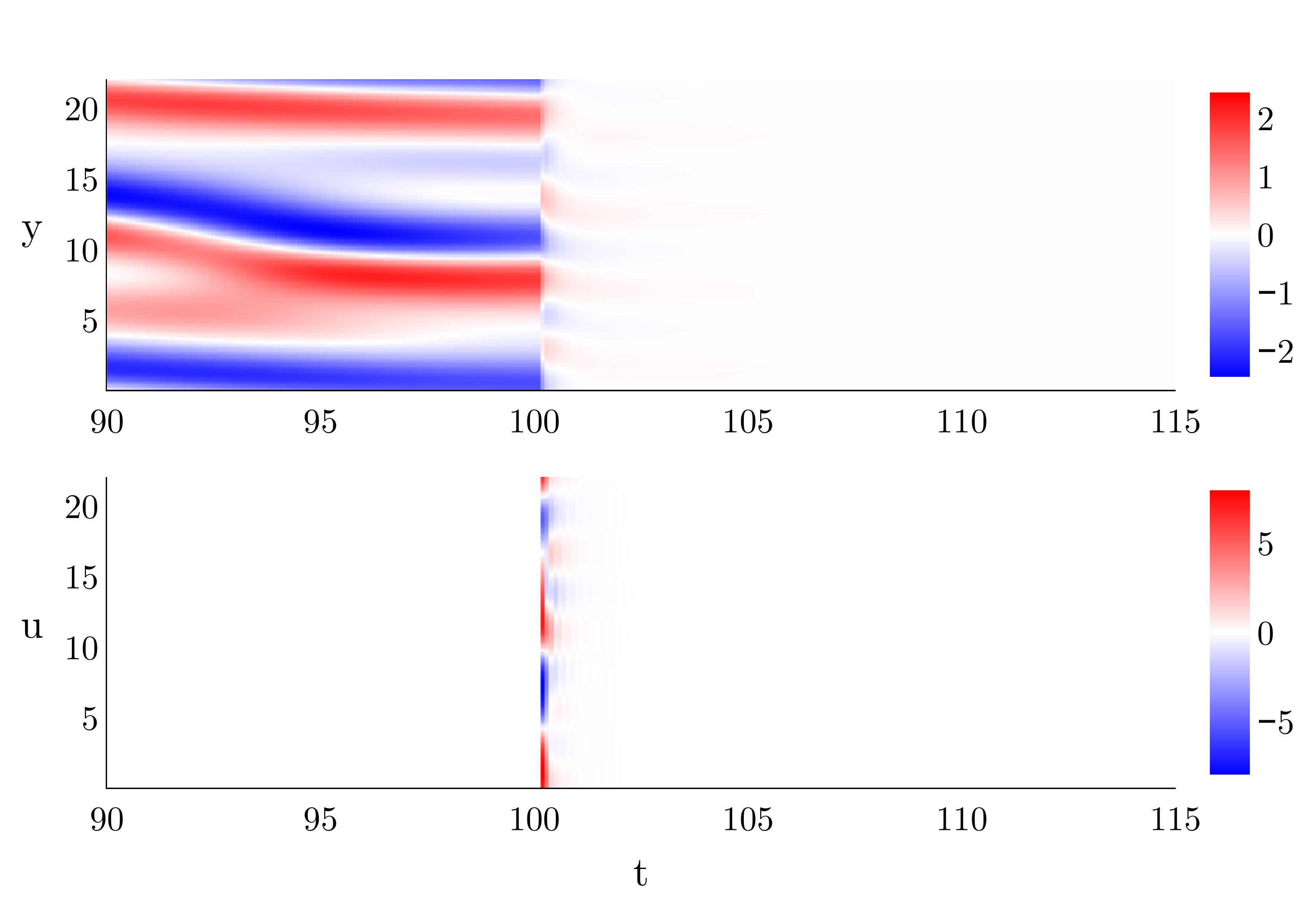}\\
	(a)
	}
	\hfill
	\parbox[b]{\columnwidth}{
	    \centering
		\includegraphics[width=\columnwidth]{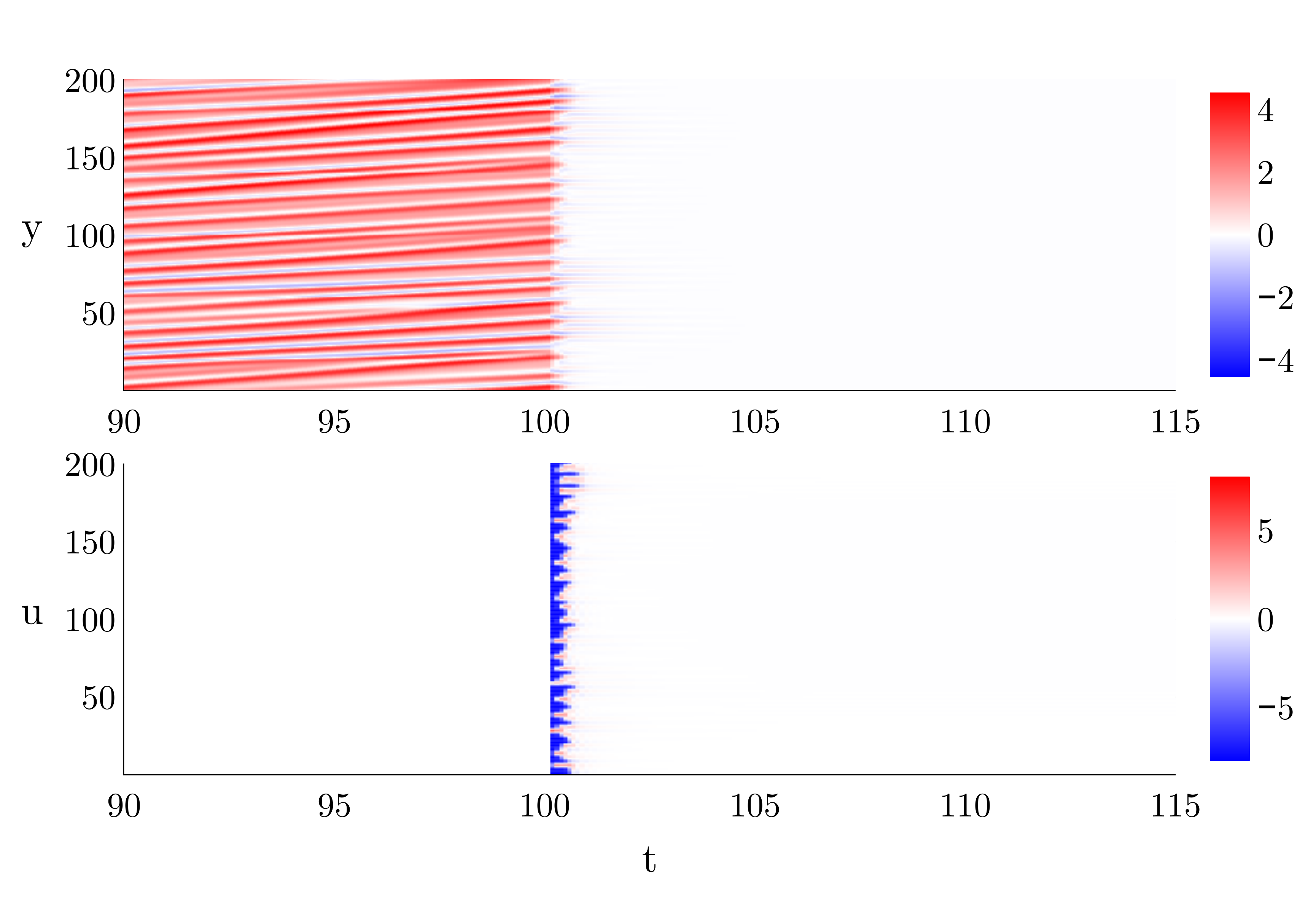}\\
		(b)
	}
 	\caption{Kuramoto--Sivashinsky. results for (a) $\mu = 0$, $L=22$ and $M=P=8$ sensors and actuators and (b) $\mu = 0.02$, $L=22$ and $M=P=8$ sensors and actuators.}
	\label{fig:KS_L22_L200_mu002}
\end{figure}

In our numerical experiments, we begin with $\mu=0$, $L=22$ (in accordance with~\cite{BSA+19,LW21,ZG21}) and set $M=P=8$, i.e., the dimension of the state and the input are identical. As the kernel $\psi$, use equidistantly placed Gaussians with $\sigma=0.8$. Eight identical RL agents are placed at these centers $c_i$. 
The respective environments consist of a single sensor, i.e., $\{\tilde{y}_{i}\}$, which means $S=1$. 
The control dimension is one, as only the action at $c_i$ has to be determined. 
For the actor and critic networks, \revision{we can thus simply use fully connected feed forward nets with a single hidden layer and a small number of neurons, respectively, cf.\ Remark \ref{rem:ComparisonCNN} as well as Table \ref{tab:KS} for details}. Due to this small dimension and the increased amount of data by a factor of $P=8$, training is very fast and takes just a few minutes on a standard desktop computer. The details of the numerical setup are also summarized in Table \ref{tab:KS}.

\begin{figure}[h]
    \centering
    \includegraphics[width=\columnwidth]{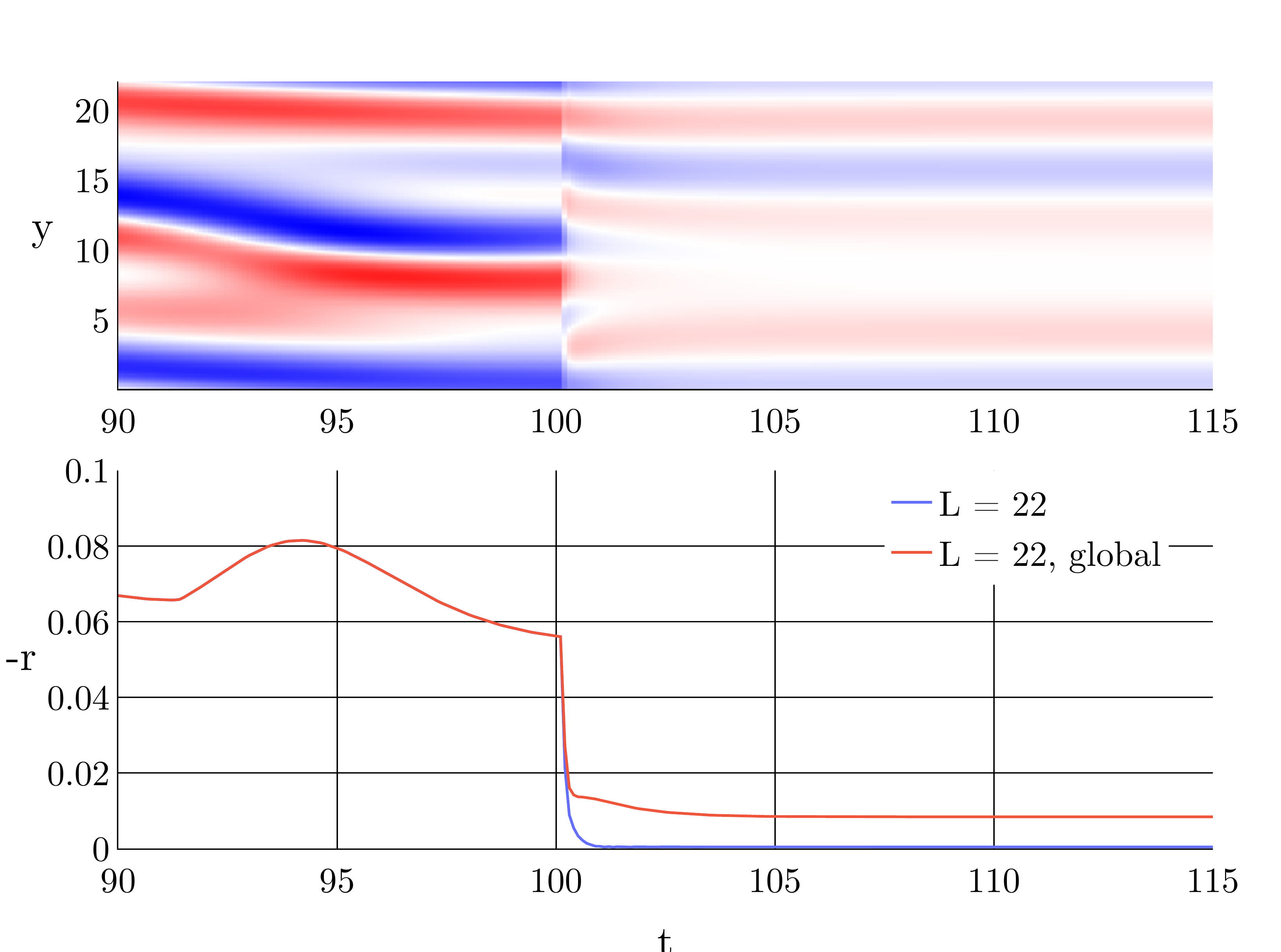}
	\caption{\revision{Performance of the fully trained global agent and it's reward curve in comparison to the curve from the previously shown experiment.}}
	\label{fig:KSglobal}
\end{figure}

The result is shown in Figure~\ref{fig:KS_L22_L200_mu002} (a). Using a random initial condition, we let the system evolve autonomously for $100$ time units before activating the agent. After that, the system is stabilized in little more than 5 seconds. This is much faster than what was achieved in~\cite{BSA+19,LW21,ZG21}, although the results are not directly comparable, as the objectives as well as the control dimensions are different. \revision{The direct comparison to a single global agent is shown in Fig.\ \ref{fig:KSglobal}. We see that the performance is inferior, even though the training time was approximately ten times higher. A scaling towards much larger domains, as we will do next, thus quickly becomes infeasible for single agents.}
%What is even more important is the fact that the standard implementation of the DDPG agent in \emph{Stable Baselines 3} \cite{stable-baselines3} was not able to solve this problem with a single, global agent with $M=P=8$. %, even after $10^6$ training steps.

%\begin{figure}[h]
% 	\centering
%    ~\hspace{-2em}\includegraphics[width=\columnwidth]{graphics/KS200_80_mu.pdf}
%	\caption{Kuramoto--Sivashinsky: results for $\mu = 0.02$, $L=200$ and $M=P=80$ sensors and actuators.}
%	\label{fig:KS_L200_mu002}
%\end{figure}
    
\begin{figure}[t!]
 	\centering
 	\parbox[b]{\columnwidth}{
    \centering
	\includegraphics[width=\columnwidth]{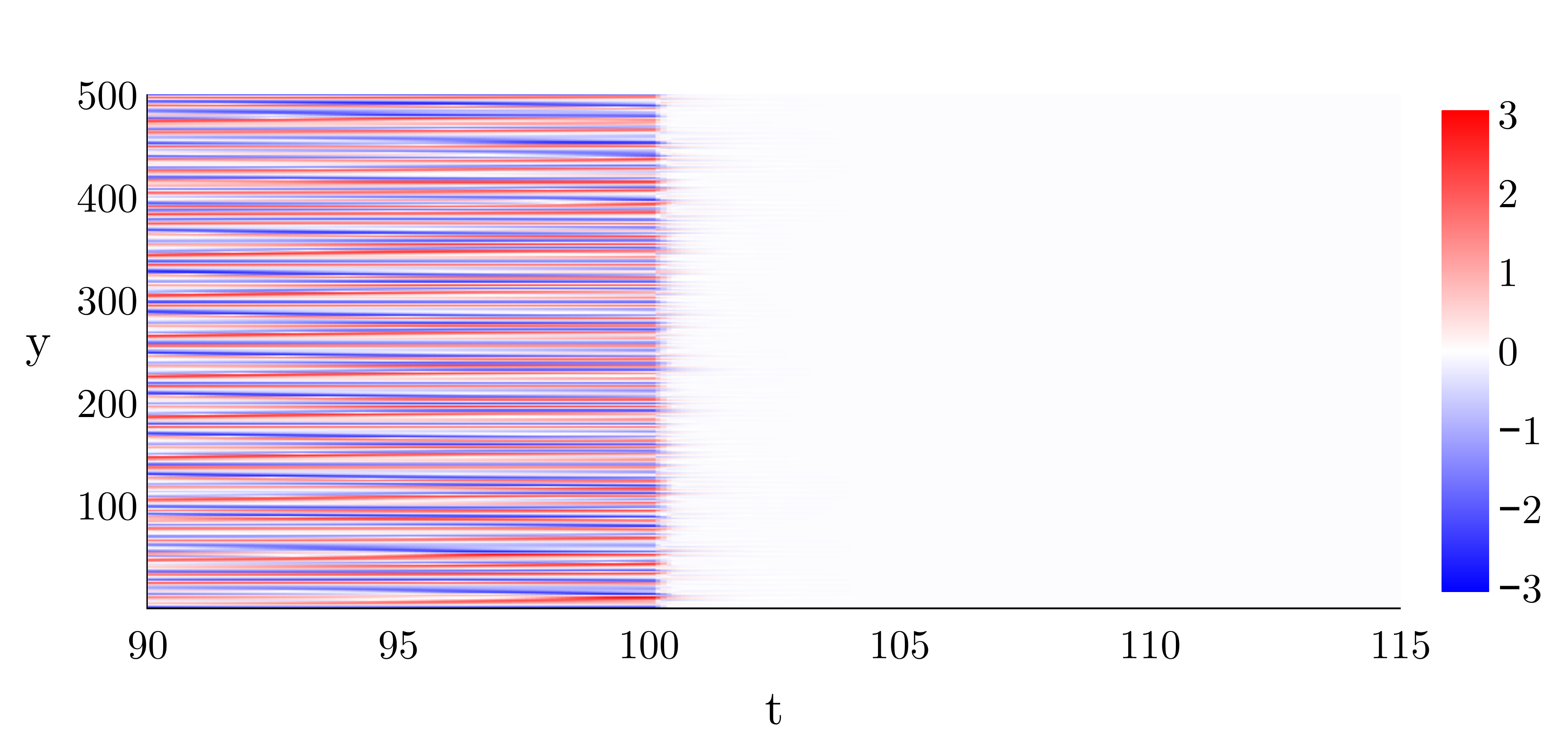}\\
	(a)
	}
	\hfill
	\parbox[b]{\columnwidth}{
    \centering
	\includegraphics[width=\columnwidth]{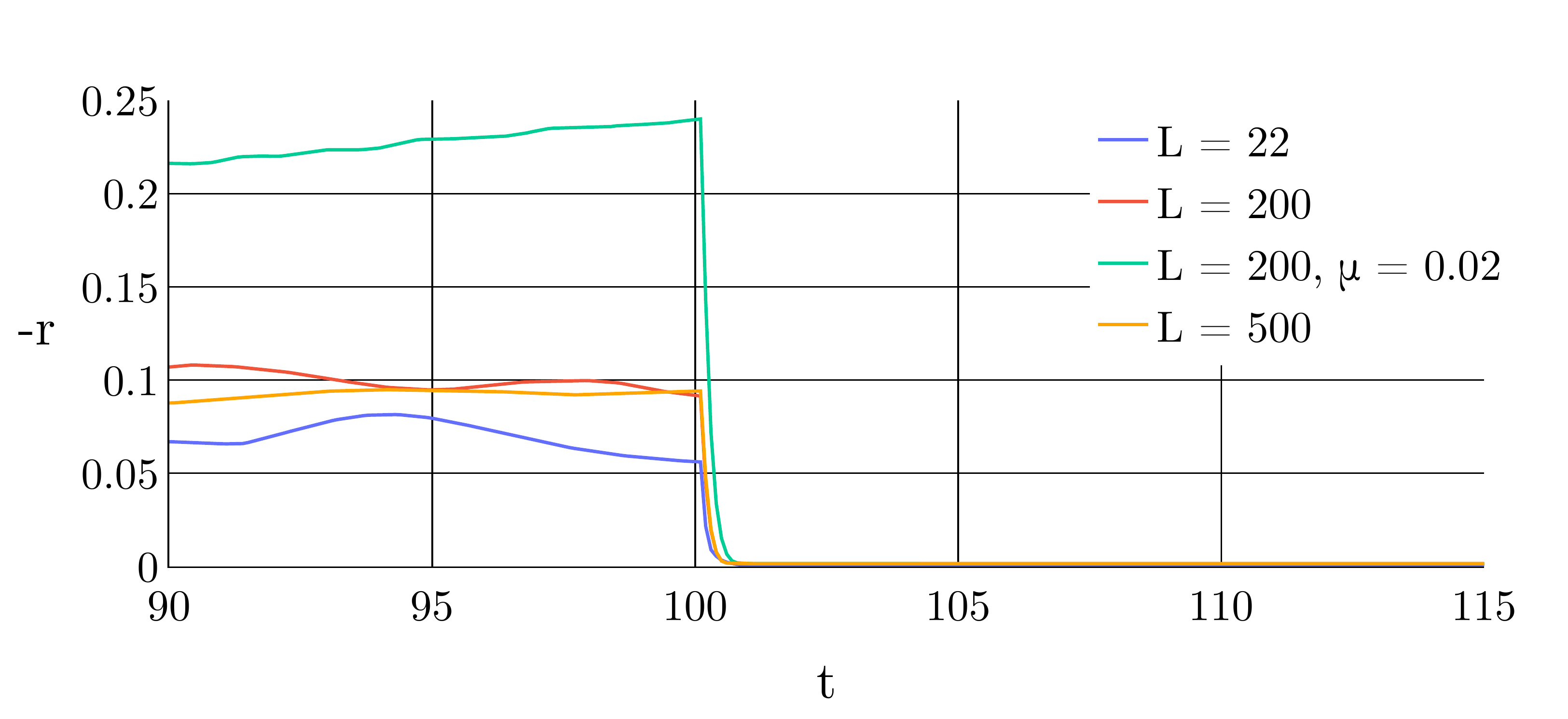}\\
	(b)
	}
 	\caption{Kuramoto--Sivashinsky. (a) Results for $\mu = 0$, $L=500$ and $M=P=200$ sensors and actuators. (b) Negated rewards (i.e., $0$ is optimal, similar to standard control formulations) over time for the conducted Kuramoto--Sivashinsky experiments. The agent starts at $t=100$.}
	\label{fig:KS_L500_rewards}
\end{figure}    

Having confirmed the validity of our approach, we now exploit the scalability and significantly extend the domain size to $L=200$ while increasing the number of sensors and actuators to $M=P=80$, i.e., $L/M=2.5$. This results in more complicated patterns; not only due to the increased domain size, but also because $L$ serves as a system parameter and the dynamics become more chaotic with increasing $L$~\cite{PHG+18,EBMR19}.
%At the same time, we even decrease the environment size in our agent to $S=3$ as well as the number of trainable parameters to two hidden layers with only 10 neurons for both the actor and the critic.
Nevertheless, we keep the low dimension of the actor and critic networks constant.
Interestingly, the performance is identical compared to the $L=22$ case. 
The reward curve is depicted in Figure \ref{fig:KS_L500_rewards} (b).

To show the robustness of convolutional sensors, we now apply the same agent to the KS system with $\mu=0.02$, i.e., we allow for a small inhomogeneous disturbance. Figure \ref{fig:KS_L22_L200_mu002} (b) demonstrates that this only has a minimal negative influence on the performance, see also the reward curves in Figure \ref{fig:KS_L500_rewards} (b). Finally, a straightforward transfer of the agent from $L=200$ to $L=500$ -- now with $M=P=200$ -- is possible just as easily. This is visualized in Figure \ref{fig:KS_L500_rewards} (a), where we have increased both $M$ and $P$ by a factor of 2.5 while leaving all other parameters unchanged. This way, the local agents have the same $L/M$ ratio and hence, we can simply use the agent trained on the smaller domain. 

\revision{It should be mentioned that the control setting is quite favorable for stabilization, as the distributed controller has a lot of authority over the system dynamics. As a matter of fact, a manually tuned opposition controller could achieve almost the same performance as our RL agent. To show that the concept also holds in more challenging situations, we will consider a two-component system next.}

\subsection{Keller-Segel model for Chemotaxis}
In our second example, we study a Keller-Segel type model for Chemotaxis~\cite{PH11}, 
which is a process that describes the movement of cells (or organisms) in response to the presence of a chemical signal substance inhomogeneously distributed in space:
\begin{equation*}%\label{eq:Chemotaxis}
    \begin{aligned}
        \pder{y}{t} &= \pder{}{x} (D\pder{y}{x} - \chi y \pder{z}{x}) + q y(1-y), \\
        \pder{z}{t} &= \pderHigher{z}{x}{2} + y - z + f(x,u),
    \end{aligned}
\end{equation*}
with homogeneous Neumann boundary conditions. Here, $y$ and $z$ denote the cell density and the chemoattractant concentration, respectively, at time $t$ and location $x\in[0,L]$.  Furthermore, $D$ is the cell diffusion coefficient, $\chi$ is the chemotactic coefficient and $q$ is the growth rate. The control task is to steer the chemoattractant $z$ (via $f$ as above) in order to stabilize the cell density $y$ at $y=1$, i.e.,
\[
    r = -\ell= -\left(\langle (y-1)^2 \rangle + \alpha \sum_{i=1}^P u_i^2 \left\langle \psi^2(x - c_i) \right\rangle\right).
\]
Optimal control of Keller-Segel type models has been studied theoretically (see, e.g.,~\cite{RY01}), but there is until now little work on numerical results; see~\cite{DP19} for an exception using a slightly different model.

\begin{figure}[h]
    \centering
    \includegraphics[width=\columnwidth]{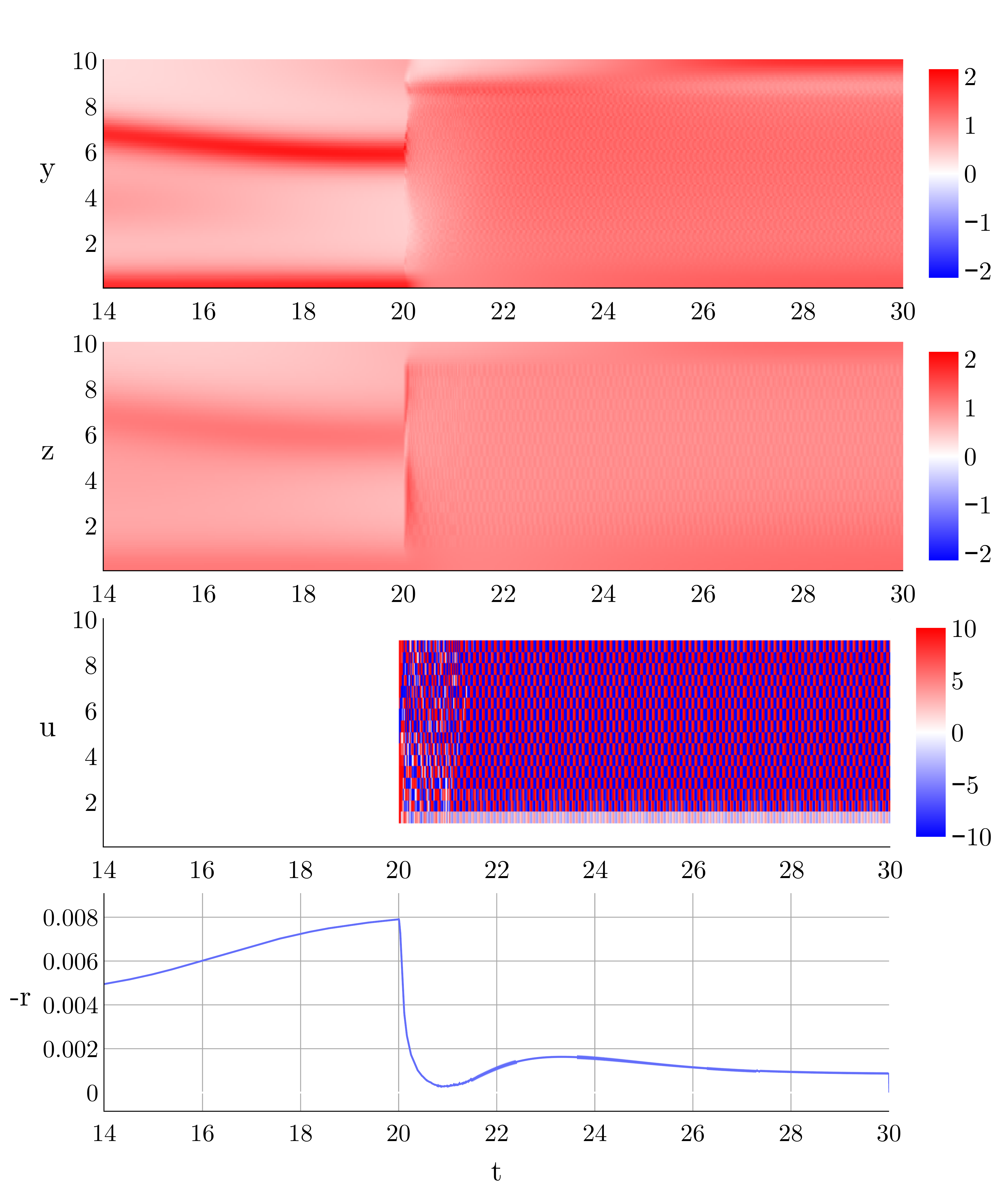}
	\caption{The components $y$ and $z$, the action $u$ and the negated reward signal $r$ averaged over the actuators for Keller-Segel. Agent starts at $t=21$.}
	\label{fig:Chemotaxis}
\end{figure}

In our numerical example, we set $D=q=1$, $\chi = 5.6$ and $L=10$.
As the convolution kernel, we use indicator functions with width $0.25$, placed equidistantly at $M=40$ locations. We again choose a local environment dimension of $S=3$ and as we do not have periodic boundary conditions, we can only place $P=36$ actuators inside the domain (with ``distance'' at least two to each boundary). As the state, we consider the convolved states $\tilde{y}$ and $\tilde{z}$ as well as a time-delayed observation of both $\tilde{y}$ and $\tilde{z}$ with $\Delta t = 1$, meaning that each agent gets $S \cdot 2 \cdot 2 = 12$ inputs. Both actor and critic again consist of a fully connected feed forward neural network with two hidden layers consisting of $20$ neurons each.
The results are shown in Figure \ref{fig:Chemotaxis}. Even though we cannot directly control $y$ but only $z$, we see that our approach performs very well in stabilizing the cell density $y$.

\subsection{2D Turbulence}
As the final and most complex case, we consider a fluid problem, more specifically two-dimensional decaying isotropic turbulence, see~\cite{TNB16} for a detailed description and the numerical solver. The dynamics are described by the two-dimensional vorticity transport equation:
\begin{align*}
    \frac{\partial \omega}{\partial t} + y_j \frac{\partial \omega}{\partial x_j} = \frac{1}{Re} \frac{\partial^2 \omega}{\partial x_i \partial x_j} + f(x,u),
\end{align*}
where $y$ and $\omega$ are the velocity and vorticity variables, respectively. The Reynolds number is defined as $Re = \frac{y^* \ell^*}{\nu}$, where $\nu$ is the kinematic viscosity. $y^*(t_0)=[\overline{y^2}(t0)]^{1/2}$ and $\ell^*(t_0)=[2\overline{y^2}(t_0)/\omega^2(t_0)]^{1/2}$ are the velocity (normalized by the square root of the spatial average of the initial kinetic energy) and the initial integral length scale, respectively.

\begin{figure}[h!]
    \centering
    \includegraphics[width=\columnwidth]{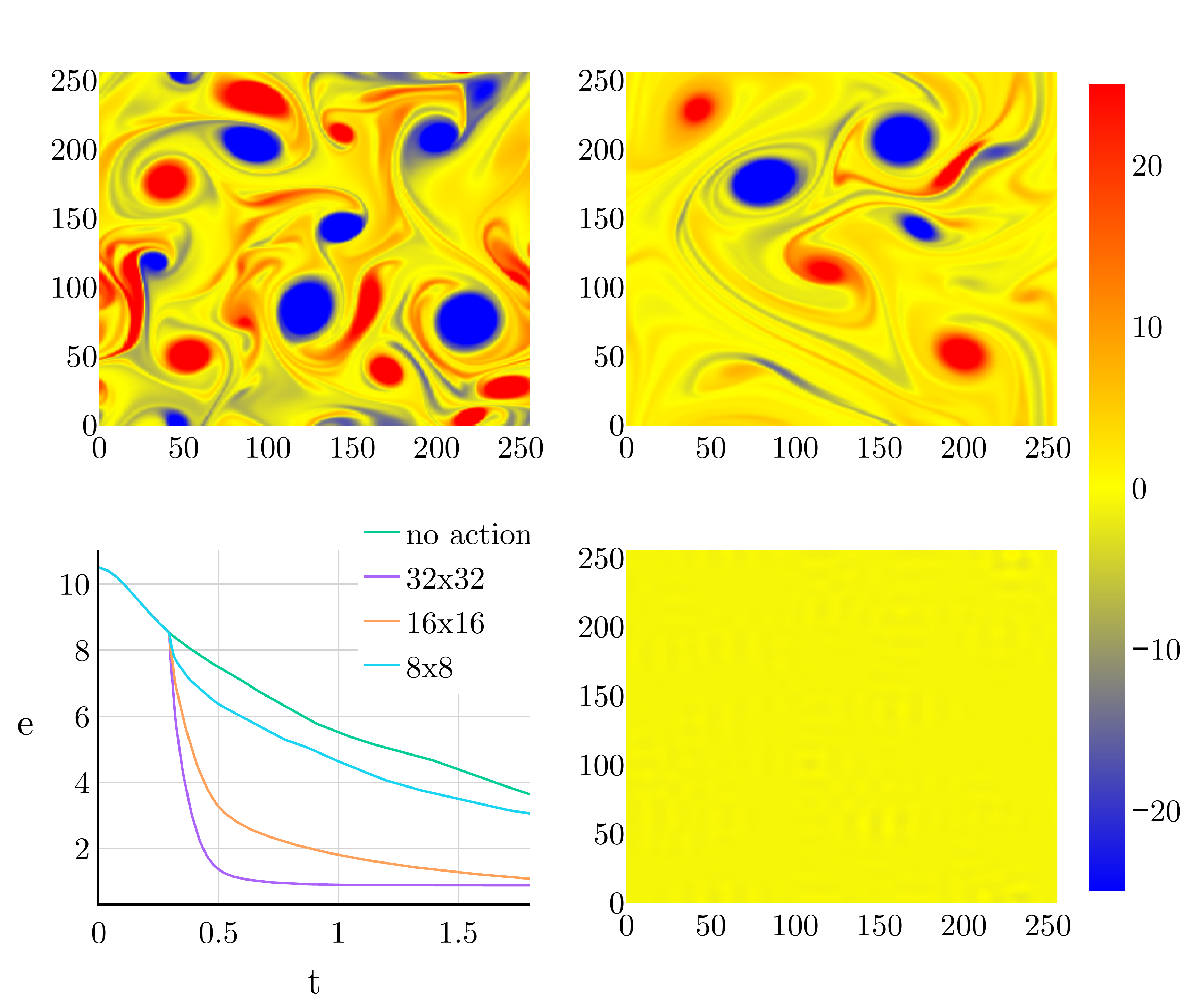}
	\caption{Convolutional RL of 2D isotropic turbulence for different numbers $M$ of identical agents (bottom left). In all experiments, the local agents' state space dimension is $S=3\times 3$. Top left: vorticity at $t=1.0$ (the time at which we activate the control). Top right: vorticity at $t=5.0$ witout control. Bottom right: the controlled solution at $t=5.0$ ($M=32\times 32$).}
	\label{fig:Turbulence}
\end{figure}

Similar to the previous two examples, our goal is to stabilize the system. Due to the viscosity term, the uncontrolled system is self-stabilizing, but a a very slow rate. Our aim is thus to significantly accelerate this stabilization. To this end, we compare different control settings, i.e., grids of $8\times 8$, $16\times 16$ and $32 \times 32$ sensors and actuators, respectively. In each test the DDPG agent has a state space of size $S = 3 \times 3$. The results are shown in Figure \ref{fig:Turbulence} in which we observe very good control performance using yet again very small neural networks (a single hidden layer with only four neurons for both the actor and critic networks).

Even though the figures show that our framework is effective in stabilizing the system, additional evaluation is called for to compare the performance the performance against a simple
%we would like to remark that a simple 
opposition control scheme. % can also do the job in this case. 
%Thus, i
Moreover, it will be interesting for future research to study systems that are more challenging to stabilize, such as the 3D channel flow~\cite{BMT01}, which has tremendous fundamental and industrial implications.

%%%%%%%%%%%%%%%%%%%%%%%%%%%%%%%%%%%%%%%%%%%%%%%%%%%%%%%%%%%%%%%%%%%%%%%%%%%%%%%%%%%%%%%%%%%%%%
\section{Conclusion}

We have shown that exploiting system knowledge, here in terms of symmetries and finite-velocity transport of information, can help us in massively reducing the complexity of reinforcement learning in distributed PDE control. Our convolutional framework allows for the ``cloning'' of many small agents with shared parameters as well as shared training data. This yields efficient control strategies for very large domains and allows for an easy transfer between different domains. In addition, the convolution operation in observing the state yields robustness against inhomogeneous disturbances.

For future work, there are several options to extend this framework. Most importantly, the question of global reward functions (even though they are likely slightly less important in physics applications) as well as control of only parts of the domain need to be addressed. Furthermore, it will be interesting to see whether additional knowledge (e.g., in the form of differential equations) can help us in further increasing the efficiency. 
In addition, it may be fruitful to study how the trained controller varies with physical parameters, such as the Reynolds number.  This may lead to parameterized control laws that are valid over larger ranges of operating conditions.  
Similarly, it may also be interesting to investigate self-similarity of the control law in the context of spatially developing turbulence, such as a developing boundary layer.  
Finally, other symmetries and invariances may be similarly embedded into the RL framework for the control of more complex spatiotemporal processes such as turbulent channel flows or Rayleigh--B{\'e}nard convection.

%%%%%%%%%%%%%%%%%%%%%%%%%%%%%%%%%%%%%%%%%%%%%%%%%%%%%%%%%%%%%%%%%%%%%%%%%%%%%%%%%%%%%%%%%%%%%%
\section*{Code}
\revision{The source code is written in the language Julia and is publicly available under \url{https://github.com/janstenner/DistributedConvRL-PDE-Control}.}

\section*{Acknowledgements}
%Will be inserted later.
SP acknowledges support by the Priority Programme 1962 of the Deutsche Forschungsgemeinschaft (DFG) and by the project ``SAIL'' (Grant ID NW21-059A) of the Ministry of Culture and Science of the State of Northrhine Westphalia, Germany. All authors from Paderborn (SP, JS, VC, OW) acknowledge support by the BMBF within the project ``DARE''. 
KT thanks the support from the US Air Force Office of Scientific Research (FA9550-21-1-0178) and the Vannevar Bush Faculty Fellowship (N00014-22-1-2798).

\bibliographystyle{elsarticle-num} 
\bibliography{literature}

%\endinput
%%
%% End of file `elsarticle-template-num.tex'.
\end{document}